\documentclass[]{interact}

\usepackage{epstopdf}
\usepackage[caption=false]{subfig}

\usepackage[numbers,sort&compress]{natbib}

\usepackage{url}
\newif\ifarxiv
\arxivtrue

\ifarxiv
\else
\usepackage{lineno}
\linenumbers
\fi

\bibpunct[, ]{[}{]}{,}{n}{,}{,}
\renewcommand\bibfont{\fontsize{10}{12}\selectfont}

\theoremstyle{plain}

\theoremstyle{definition}

\theoremstyle{remark}

\begin{document}

\articletype{RESEARCH ARTICLE}

\title{Automatic detection of faults in race walking from a smartphone camera: a comparison of an Olympic medalist and university athletes}

\ifarxiv
\author{\name{Tomohiro Suzuki\textsuperscript{a}, Kazuya Takeda\textsuperscript{a}, and Keisuke Fujii\textsuperscript{*a,b,c}\thanks{*Corresponding author: Keisuke Fujii. Email: fujii@i.nagoya-u.ac.jp}}
\affil{\textsuperscript{a}Nagoya University, Furo-cho, Chikusa, Nagoya, Aichi, Japan; \textsuperscript{b}RIKEN Center for Advanced Intelligence Project, 744 Motooka, Nishi, Fukuoka, Japan; \textsuperscript{c}JST PRESTO, 7 Gobancho, Chiyoda, Tokyo, Japan}
}
\else
\author{Anonymous}
\fi
\maketitle

\begin{abstract}
Automatic fault detection is a major challenge in many sports. 
In race walking, referees visually judge faults according to the rules. Hence, ensuring objectivity and fairness while judging is important. To address this issue, some studies have attempted to use sensors and machine learning to automatically detect faults. However, there are problems associated with sensor attachments and equipment such as a high-speed camera, which conflict with the visual judgement of referees, and the interpretability of the fault detection models. In this study, we proposed a fault detection system for non-contact measurement. We used pose estimation and machine learning models trained based on the judgements of multiple qualified referees to realize fair fault judgement. We verified them using smartphone videos of normal race walking and walking with intentional faults in several athletes including the medalist of the Tokyo Olympics. The validation results show that the proposed system detected faults with an average accuracy of over $90\%$. We also revealed that the machine learning model detects faults according to the rules of race walking. In addition, the intentional faulty walking movement of the medalist was different from that of university walkers. This finding informs realization of a more general fault detection model.
\ifarxiv
The code and data are available at \url{https://github.com/SZucchini/racewalk-aijudge}.
\fi
\end{abstract}


\section{Introduction}
Race walking is one of the long-distance track and field Olympic events, which has been also studied as biomechanical research \cite{hoga2017reconstruction, pavei2016effects, gomez2018race}. There are two rules in race walking: at least one foot must always be in contact with the ground and the forward knee must be extended until it is perpendicular to the ground \cite{worldathletics2022rule}. Violation of the first rule is called \textit{loss of contact} (LC) violation and that of the second rule is called \textit{bent knee} (BK) violation. The walker who commits these violations will be given a warning and disqualified after three warnings.

Referees in race walking check such faults visually \cite{worldathletics2022rule}. Then, the standard of judgement varies depending on the ability of the referees and the type of the competition (e.g., record meeting or championship). In addition, because referees check several walkers at the same time, they may not be able to fully observe all walkers. For spectators who are not familiar with race walking, it is difficult to understand whether a fault has been committed and to appreciate the sport. Therefore, it is important to realize objective and fair judgement. It is also necessary to clarify the reasons for the judgements and devise ways to make it easier for spectators to understand race walking.

To achieve objective and fair judgements, some approaches detected LC using piezoelectric sensors on the plantar feet \cite{santoso2013development} or acceleration sensors attached to the waist and lower limbs \cite{lee2013detection, di2016outdoor}.
Another study detected two faults using machine learning \cite{taborri2019automatic}. However, all of these studies used sensors attached to the body, which is not practical because they may affect the performance of walkers. In addition, some machine learning methods (e.g., support vector machine \cite{vapnik1999nature} used in \cite{taborri2019automatic}) may make it difficult to interpret the reasons for the fault detection and sometimes be in conflict with the visual judgement of race-walk referees.
Other sports use image recognition technology to solve similar problems. Examples include scoring in rhythmic gymnastics \cite{diaz2014automatic} and figure skating \cite{xu2019learning} and detecting offsides in soccer \cite{uchida2021automated}. These studies use competition videos and do not require sensors. Similarly, in judging the race walking competition, automatic non-contact fault detection from the walking video would contribute to reducing the burden on referees and improving the objectivity of judgement.

In this study, we propose a fault detection system using smartphone camera video to realize non-contact and objective fault detection (an overview of the proposed system is shown in Supplementary Figure 1). This study explores the feasibility of non-contact fault detection through validation of the proposed system using intentional faulty walking videos. The system first estimates key points (joint locations) from walking videos by pose estimation. We used a pose estimation model whose performance was improved by the fine-tuning technique described below. Next, the input features of the fault detection model are calculated from the coordinate data estimated by the pose estimation. Finally, the feature vector is input to the classifier, which outputs the detection results. A smartphone camera is used to capture video to simplify the system, which can be used in a wide range of situations such as practice, competition, and referee training. The effectiveness of the proposed system was verified by intentional faulty walking videos of a university student race walker and a Tokyo Olympics medalist. The contributions of this study are as follows: (1) to develop the interpretable automatic fault detection system of race walking using machine learning with non-contact measurement; (2) to improve pose estimation performance by fine-tuning the pre-trained model using images of race walking; (3) to clarify that the system can detect the fault of university race walkers with high accuracy; and (4) to clarify that an Olympic race walker and university race walkers had different movements during intentional faulty walking. 

\section{Materials and Methods}
\subsection{Experiment}
\subsubsection{Participants}
In this study, the participants were one 25-year-old Tokyo Olympics medalist (walker A) with 11 years of experience and a personal best of $38'57"37$ in the 10,000 m walk and four university track and field athletes ($21 \pm 0.8$ years old; walkers B, C, D, and E) with $5.5 \pm 0.5$ years of experience and a personal best of $47'20"37 \pm 1'19"55$ in the 10,000 m walk. The participants were fully informed about the study and their consent was obtained in advance. All the experimental procedures were performed after obtaining prior approval for ``Experiments on Human Subjects'' 
\ifarxiv
from the Graduate School of Informatics, Nagoya University.
\else
in the organization.
\fi

\subsubsection{Procedures}
The experiment was conducted at the 
\ifarxiv
Nagoya 
\fi
University track and field ground (under the conditions shown in Supplementary Figure 2). The walking section was approximately 20 m long, and we captured the side profiles of the participants during the experiment. The participants walked repeatedly with breaks in between. The participants were instructed to randomly walk with a normal form, a BK, and LC. Since walker E could not walk with a BK, he was instructed to walk with the normal form and LC. The other participants frequently walked with a BK. The experiment was conducted over multiple dates to ensure that fatigue did not affect the walk.

\subsubsection{Data Collection}
Walking videos were captured by an iPhone 11 Pro camera at 60 fps. After capturing the video, we asked three referees (One person has an International Race Walking Judge qualification, the other two have a Japan Race Walking Judge qualification) to judge whether faults were committed during the experiments by the participants. We annotated the videos according to the majority vote of the three referees. The judgement results were used as the correct labels when training the machine learning model. We captured 268 normal walking videos, 265 BK videos, and 268 LC videos. The number of videos and valid data by walker are shown in Supplementary Table 1.


\subsection{Proposed System}
An overview of the proposed system is shown in Supplementary Figure 1.
The code and data are available at \url{https://github.com/SZucchini/racewalk-aijudge}.
\subsubsection{Pose Estimation}
There are two types of pose estimation models: a top-down model that estimates the key points separately after estimating the positions of all persons in the image, and a bottom-up model that estimates all key points in the image and then groups them for each person. In this study, we used HigherHRNet \cite{cheng2020higherhrnet}, a bottom-up pose estimation model. The performance of the bottom-up model is superior to that of the top-down model in multiple-person estimation. Since the proposed system is meant to be applied to racing, the performance of multi-person estimation is important. Our system estimates 17 key points: nose, left and right eyes, ears, shoulders, elbows, wrists, hips, knees, and ankles. The input feature for the fault detection model is computed from pose estimation data, indicating that the performance of the pose estimation model affects the accuracy of the system. Therefore, we fine-tuned a pre-trained HigherHRNet model by race walking images to improve estimation performance.

\subsubsection{Fine-tuning of the Pose Estimation Model}
Fine-tuning is a learning method in which a model is transferred to another domain by updating the weights of a pre-trained model using other training data. In track and field pose estimation, fine-tuning has been used to improve the estimation performance of long jump and triple jump movements \cite{ludwig2021self}. We fine-tuned a pre-trained HigherHRNet based on common human pose images using race walking images to improve its estimation performance for walking movement. We adopted fine-tuning to reduce the cost of annotating images, aiming for a simpler system that can be used by a large number of people.

\subsubsection{Post-processing of Pose Estimation Data}
The key point coordinate data obtained by pose estimation must be corrected for the effects of differences in the walker's physique. Therefore, all data were normalized so that the length from the nose to the hip key point was 1. In addition, all points were translated so that the coordinates of the nose were set to $(0,0)$.

Next, the key point coordinates of the shank and the knee angle were calculated from the key point coordinates of the hip, knee, and ankle. The shank position was defined as the midpoint of the knee and ankle, and the knee angle was defined as $\theta$ counterclockwise from the thigh to the calf. After the knee angle was calculated, outliers were identified based on the standard deviation of the frame-to-frame change in the right knee angle for all data, and the outlier data were removed.

After removing the outlier data, we extracted the interval of ``two steps from the most bent right knee'' from each data set to unify the walking scene of the input data. The definition of the knee angle and an example of a walking scene are shown in Supplementary Figure 3. The number of frames for all walking data was normalized into 85 frames by interpolation to clarify the frames that contribute to the fault detection.

\subsubsection{Classifier}
A logistic regression model was used as the classifier for fault detection. The logistic regression model is one of the simplest machine learning models, and it can analyze the detection criteria from feature importance based on the standard regression coefficients. Since it is undesirable not to understand the process of judging sports, we use the model that allows analysis of judging criteria. The detection model was created separately for two faults (BK and LC). The model takes the key points coordinates and knee angles as input, and outputs the results of the fault detection.

\subsection{System Verification}
\subsubsection{Pose Estimation Model}
Some walking images extracted from the collected videos were used to fine-tune the pose estimation model. The walking images were annotated by the COCO Annotator \cite{cocoannotator}. We used 108 images of walkers A, D, and E as training data, 25 images of walker B as validation data, and 52 images of walker C as test data. We used MMPose \cite{mmpose2020}, an open-source framework for pose estimation, for the fine-tuning.

The average Precision (AP) and time-series changes in knee angle were used to evaluate the pose estimation models. Representative pose estimation and object detection models, including the HigherHRNet used in this study, such as Openpose \cite{openpose} and Detectron2 \cite{wu2019detectron2}, use AP for evaluation. AP is calculated from object keypoint similarity (OKS), which represents the degree of proximity between the correct and estimated points at each key point. OKS is defined by the following equation:
\begin{align}
    OKS = \frac{\sum_{i}\exp(-d_{i}^{2}/2s^{2}k_{i}^{2})\delta(v_{i}>0)}{\sum_{i}\delta(v_{i}>0)},
\end{align}
where $d_{i}$ is the Euclidean distance between the detected keypoint and the corresponding ground truth, $v_{i}$ is the visibility flag of the ground truth, $s$ is the object scale, and $k_{i}$ is a per-keypoint constant that controls falloff.


In the AP calculation, a threshold value is defined, and the average precision of all key points is calculated, assuming that the OKS of key point $i$ exceeds the threshold value as the correct answer. For general evaluation, the average value of the AP for each threshold value is used when the threshold value is changed in 10 steps of 0.05 from 0.50 to 0.95. The time-series variation in the knee angle allows for the identification of angle outliers. Because outliers are caused by incorrect key point estimation, the occurrence of incorrect estimation can be visually determined by checking the time-series variation of the knee angle.

\subsubsection{Fault Detection Model}
The fault detection model was evaluated by dividing the data by walker and cross-validating. In this method, the training walker data does not include the test data, thus the model performance for unknown walkers can be evaluated. We used the accuracy and F-score to evaluate the model performance. The accuracy is the percentage of the estimated results that are correct. The F-score is expressed as F-score = (2 $\times$ Precision $\times$ Recall) / (Precision + Recall), where the Recall is equal to the true-positive rate, and the Precision is defined as the ratio of the sum of true positives and true negatives to false positives. The BK detection model was evaluated using data from walkers A to D. The LC detection model was evaluated using data from walkers A to E.

To determine why faults were detected, we used the standard regression coefficient for the logistic regression model as a measure of feature importance. In the analysis of feature importance by standard regression coefficients, the input features with larger absolute values of the coefficients contribute more to the detection results. We classified the input features into nine categories: features related to the x and y coordinates of the hip, knee, shank, and ankle, and features related to the knee angle. For the analysis of the detection reason, we used the average of the absolute values of the standard regression coefficients of the four (BK detection) or five (LC detection) models created for cross-validation.

\section{Results}
\begin{figure}[b]  
    \begin{center}
    \includegraphics[width=\linewidth]{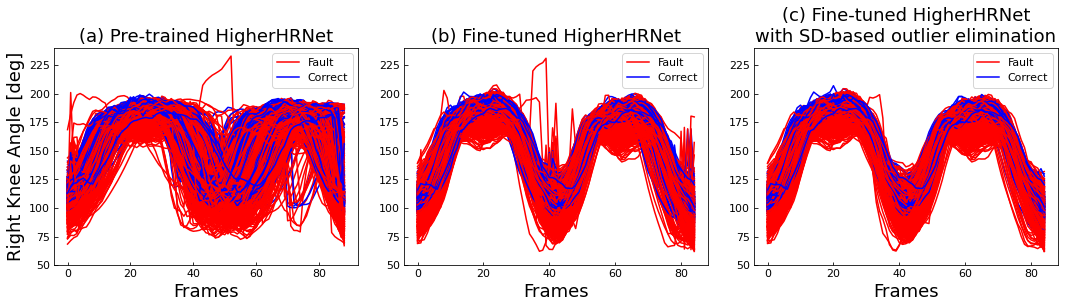}	
    \caption{Comparison of time-series changes in the right knee angle to all data. (a) Knee angles calculated from pre-trained HigherHRNet results; (b) knee angles calculated from fine-tuned HigherHRNet results; and (c) knee angles calculated from fine-tuned HigherHRNet results with SD-based outlier elimination} 
    \label{fig:res_alldata}
    \end{center}
    \vspace{-20pt}
\end{figure}

\subsection{Pose Estimation Performance}
First, we show the results of the pose estimation performance. 
AP increased by 0.031 from 0.961 to 0.992 before and after fine-tuning. Figure \ref{fig:res_alldata} compares the time-series change in the right knee angle of all data by the pose estimation model. These results reveal that outliers in the angle are reduced in the estimation using HigherHRNet (Fine-tuning), indicating that the fine-tuning improves the estimation performance. The application of outlier removal also smoothes out the angle changes. We also show the time-series change in the right knee angle of the test data by the estimation model in Supplementary Figure 4.

\subsection{Fault Detection Model Performance}
Table \ref{tab:model_eval} shows the results of the fault detection performance for the input data generated from the HigherHRNet (fine-tuning). The average accuracy was over $90\%$. However, the detection of faults in the video of walker A tended to be lower than that of the other walkers. The LC detection result for walker A was worse than the BK result.

    \begin{table}[h]
        \centering
        \caption{Fault detection model evaluation for each walker and fault.}
        \vspace{5pt}
        \begin{tabular}{lrrrr}
            \hline
            & \multicolumn{2}{c}{Bent Knee} & \multicolumn{2}{c}{Loss of Contact} \\
            & \multicolumn{1}{c}{Accuracy} & \multicolumn{1}{c}{F-score} & \multicolumn{1}{c}{Accuracy} & \multicolumn{1}{c}{F-score} \\
            \hline
            A & 0.850 & 0.893 & 0.700 & 0.727\\
            B & 0.966 & 0.961 & 0.960 & 0.959\\
            C & 0.915 & 0.905 & 0.967 & 0.967\\
            D & 0.972 & 0.966 & 1.000 & 1.000\\
            E & - & - & 0.969 & 0.969\\
            \hline
        \end{tabular}
        \label{tab:model_eval}
    \end{table}

\section{Discussion}
In this study, a system for detecting faults from walking videos was constructed for the purpose of realizing non-contact fault judgement in race walking. Here, we first discuss the results of fine-tuning the pose estimation model and then analyze the feature importance of the model to clarify the reason for fault detection. We also discuss the causes of the poor fault detection performance for the movements of walker A. In addition, we discuss the movement differences between the training and test data may cause failure of fault detection. Finally, we discuss the future potential and challenges of our system.

Typical pose estimation models are trained on large amounts of data, such as the COCO dataset \cite{lin2014microsoft}. In this study, we were able to improve the performance of a pre-trained model by fine-tuning it with approximately 100 race walking images. The training was successful even with a small amount of data because the race walking movement consists of repetitions of specific periodic movements. 
In this sense, this method may also be effective for running and sprinting. 

\begin{figure}[t!]  
    \begin{center}
    \includegraphics[width=\linewidth]{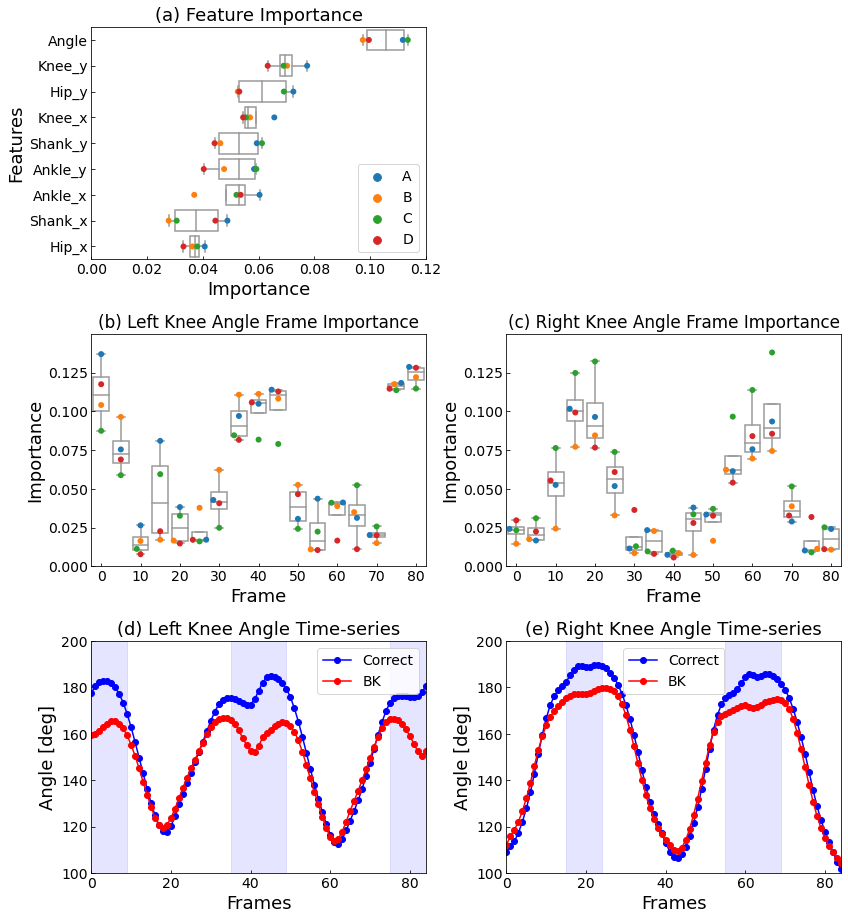}	
    \caption{Feature importance analysis of BK. (a) Feature importance of the BK detection model. (b)–(c) Knee angle's frame importance of the BK detection model. (d)–(e) Comparison of time series changes in the knee angle between normal and BK walking (The frames with relatively high feature importance are shown in light blue).} 
    \label{fig:bk_all}
    \end{center}
    \vspace{-20pt}
\end{figure}

We show the feature importance of the BK detection model in Figure \ref{fig:bk_all}(a). This figure reveals that the knee angle is of utmost importance in BK detection.
Next, we analyzed the importance by frame of the input feature of knee angle. Figures \ref{fig:bk_all}(b) and \ref{fig:bk_all}(c) show the average importance for each of the five frames for the left and right knee angles. Frames 0–9, 35–49, and 75–84 had high importance for the left knee angle, while frames 15–24 and 55–69 had high importance for the right knee angle.
From the above results, the BK detection of the model was considered to be based on the knee angle in specific frames. Figures \ref{fig:bk_all}(d)–\ref{fig:bk_all}(e) shows the average time-series change in knee angle for all walkers during normal walking and BK walking. The frames with high feature importance indicate the pose from when the front foot touches the ground until it becomes vertical. In BK walking, the knee at this time is more bent than in normal walking. An example image of the pose is shown in Supplementary Figure 6(a). Therefore, the BK detection of the model would follow the rules of race walking (e.g., the timing and the knee angle).

In LC detection, the importance of the input features is shown in Supplementary Figure 5(a), which shows that the importance of the y-coordinate of each key points is high for LC detection. The importance of the knee angle is also relatively high, but this will be discussed later.
Next, we analyzed the feature importance of the knee key point's y-coordinate by frame, which is one of the most important features. Supplementary Figures 5(b) and 5(c) show the average importance for each of the five frames for the left and right knee key point's y-coordinate. There is no significant difference in importance between frames for the left side, while frames 45–49 and 70–74 are more important for the right side.

From the above results, we considered that the LC detection model focuses on the y-coordinate of each key point to detect the LC. Supplementary Figure 5(e) shows the average time-series change in the y-coordinate of the right knee for all walkers during normal walking and LC walking. In the important frames, especially in frames 45–49, the knee position is higher in LC walking than in normal walking. The left knee y-coordinate shown in Supplementary Figure 5(d) also differed between normal and LC walking. In the relevant frames, the walker is swinging his feet forward, a situation where both feet may be off the ground. An example image of the pose is shown in Supplementary Figure 6(b). Therefore, the LC detection of the model would follow the rules of race walking (e.g., the timing and the height of each joint).

\begin{figure}[t]  
    \begin{center}
    \includegraphics[width=\linewidth]{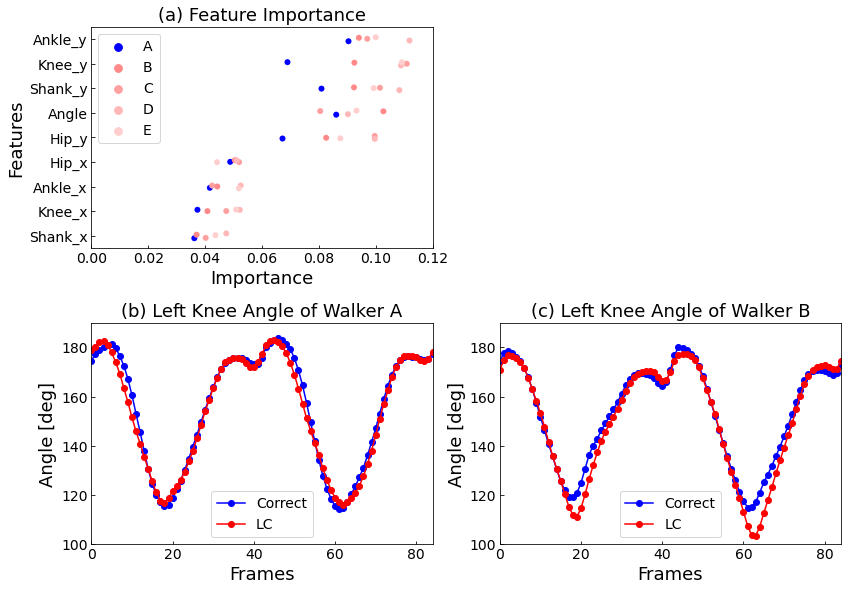}	
    \caption{Comparison of walker A and other walkers in LC. (a) Feature importance comparison of the detection models for walker A and other walkers (we removed boxplots because of the comparison). (b)–(c)Left knee angle of walker A and walker B}
    \label{fig:lc_comp_all}
    \end{center}
    \vspace{-10pt}
\end{figure}

The model could not detect LC in the movements of walker A well. Moreover, the knee angle, which according to the rules should not be the focus of attention during LC detection, became more important. We determined cause of these results by comparing the movements of walker A and other walkers.
First, to clarify the features of the detection model of walker A (an Olympian), we compared the feature importance of the detection model of walker A (the model trained on data other than walker A's) with the average feature importance of the detection models of the other walkers. Figure \ref{fig:lc_comp_all}(a)(b) shows that the feature importance of the knee angle was the fifth highest in the detection model for all walkers except walker A, while that for walker A was the second-highest in the detection model. These results suggest that there is a difference in knee angle between normal and LC walking in the training data of the detection model of walker A (data of walkers other than walker A), whereas there is no difference in the data of walker A.

To clarify a difference in the knee angle between walker A and the other walkers, we analyzed the average time-series change in the left knee angle of walkers A and B. Figures \ref{fig:lc_comp_all}(c)(d) show that there was almost no difference in the knee angle between normal and LC walking for walker A, but there was a difference between normal and LC walking for walker B. Walker B rolled up his hind leg during LC walking, and as a result, his knee was bent more than in normal walking. Walkers C through E showed the same tendency as walker B. These results indicate that walkers B to E tended to roll up their hind legs when they intentionally walked in LC walking, while walker A did not. This difference may explain why the detection model of walker A could not use the features that were focused on during the learning process, and the detection of faults in the movements of walker A was poor. Differences in the ability to control walking movements based on athletic experience and athletic ability might affect the differences in leg roll-up movements during intentional faulty walking.

In detecting the faults in the movements of walker A, the features in the training data did not match the rules, resulting in poor performance. In this case, even if both feet were off the ground, the model could failure to detect LC when walkers do not roll up their legs. In other words, the model may have select the features that do not match the rules. We also need to take care to avoid bias in training data when building a fault detection model using machine learning. In addition, additional validation with large-scale data is necessary to clarify what kind of data is needed to create a more general detection model and accurate detection criteria.

In conclusion, the fault detection system proposed in this study achieved highly accurate non-contact fault detection under the constraint of using intentional faulty walking video data. We also showed that the reason for the detection of a fault can be analyzed by feature importance analysis of the machine learning model. If a system is created by collecting faulty walking videos during a race, a judgement assistance system for a single walker could be relatively easy to implement. On the other hand, applying the system to a race is more difficult because it is necessary to improve the performance of pose estimation for multiple persons and to solve the problem of occlusion between persons. To realize a system that can be used in races, the following issues must be addressed: (1) to collect as many natural faulty walking videos as possible, (2) to verify the versatility of the system using the collected videos, and (3) to achieve high-performance pose estimation for multiple persons.

\ifarxiv
\section{Acknowledgments}
This work was supported by JSPS KAKENHI (Grant Numbers 20H04075) and JST PRESTO (JPMJPR20CA).
\fi 

\bibliographystyle{tfnlm}
\ifarxiv

\else
\bibliography{reference}
\fi


\ifarxiv
\newif\ifarxiv
\arxivtrue

\ifarxiv
\renewcommand{\thesection}{\Alph{section}}
\setcounter{section}{0}
\setcounter{figure}{0}
\setcounter{table}{0}
\else

\documentclass[]{interact}

\usepackage{epstopdf}
\usepackage[caption=false]{subfig}

\usepackage[numbers,sort&compress]{natbib}
\usepackage{lineno}
\linenumbers
\bibpunct[, ]{[}{]}{,}{n}{,}{,}
\renewcommand\bibfont{\fontsize{10}{12}\selectfont}

\theoremstyle{plain}
\newtheorem{theorem}{Theorem}[section]
\newtheorem{lemma}[theorem]{Lemma}
\newtheorem{corollary}[theorem]{Corollary}
\newtheorem{proposition}[theorem]{Proposition}

\theoremstyle{definition}
\newtheorem{definition}[theorem]{Definition}
\newtheorem{example}[theorem]{Example}

\theoremstyle{remark}
\newtheorem{remark}{Remark}
\newtheorem{notation}{Notation}

\usepackage{lineno}
\linenumbers
\begin{document}
\fi
\section*{}
\vspace{20mm}
\Large{\bf{Supplementary materials for: \\
\\

\noindent Automatic detection of faults in race walking from a smartphone camera: a comparison of an Olympic medalist and university athletes} }

\vspace{10mm}
\ifarxiv
\noindent\large{Tomohiro Suzuki, Kazuya Takeda, Keisuke Fujii}
\else
\noindent\large{Anonymous}
\fi
\newpage
\section{Supplementary Figures}

\begin{figure}[h]  
    \begin{center}
    \includegraphics[width=110mm]{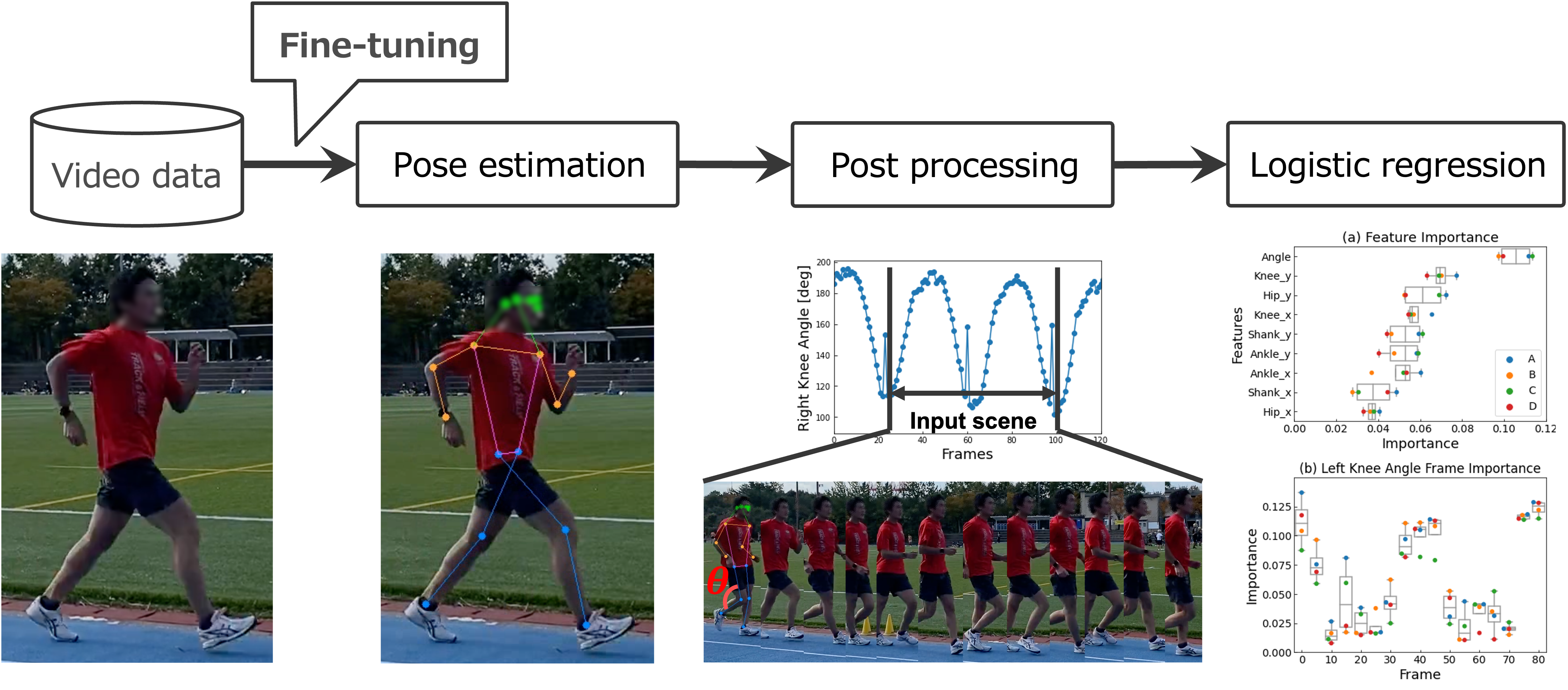}
    \caption{Overview of the proposed system} 
    \label{fig:proposed_system}
    \end{center}
\end{figure}

As illustrated in Supplementary Figure \ref{fig:proposed_system}, the proposed system consists of video input, pose estimation, data post-processing, and a classifier (judgment model).

\begin{figure}[h]  
    \begin{center}
    \vspace{30pt}
    \includegraphics[width=80mm]{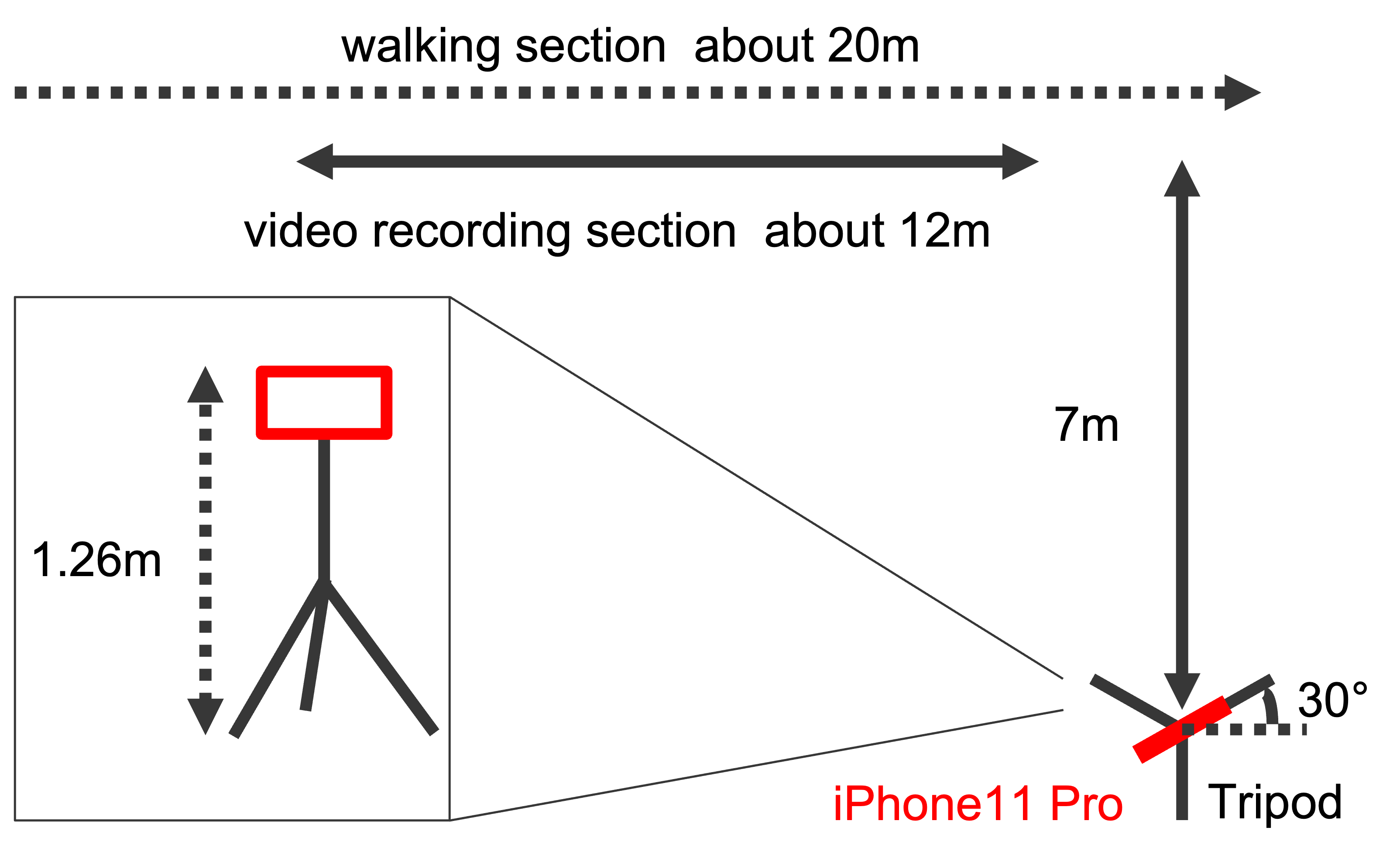}	
    \caption{Video collection experimental environment} 
    \label{fig:environment}
    \end{center}
\end{figure}

The experimental setup is illustrated in Supplementary Figure \ref{fig:environment}. We used iPhone 11 Pro to take videos. The participants' walking section was about 30 meters and the recording section was about 12 meters. The walking pace was instructed by their own race pace.

\newpage

\begin{figure}[h]  
    \begin{center}
    \includegraphics[width=100mm]{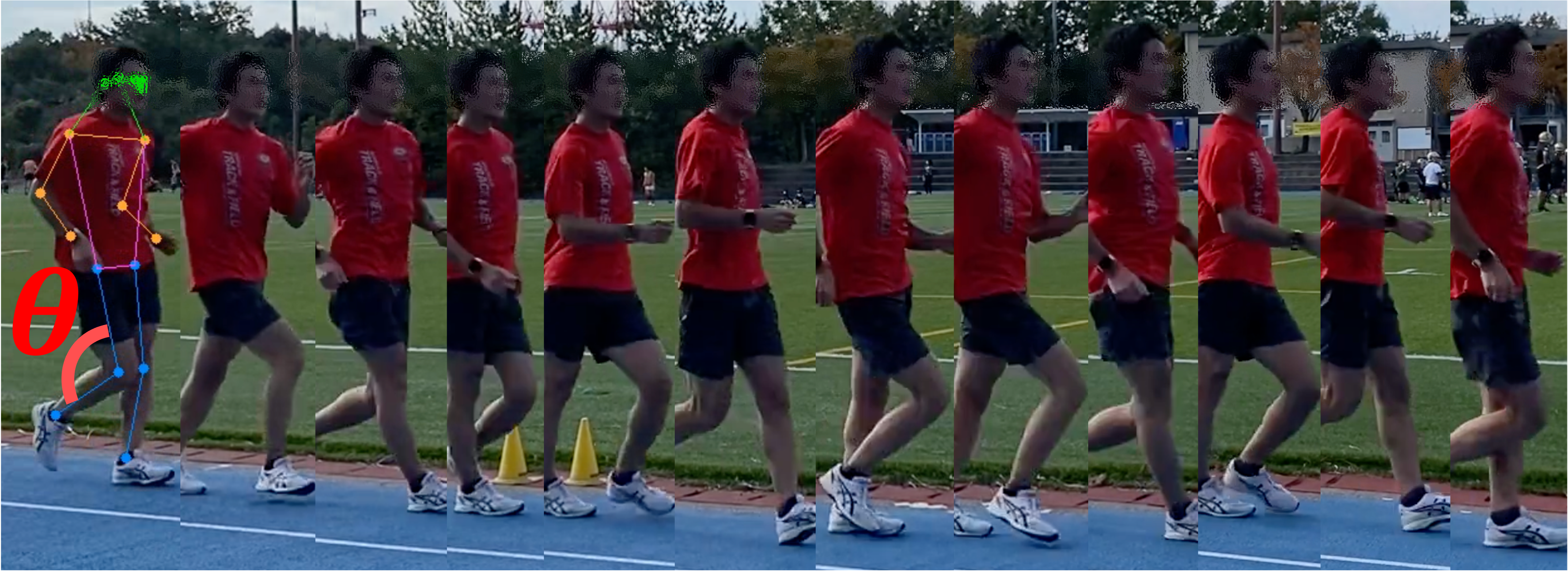}	
    \caption{The definition of the knee angle and the sequential images of the walking scene} 
    \label{fig:continuous}
    \end{center}
\end{figure}

As shown in Supplementary Figure \ref{fig:continuous}, the walking scene in the input data is a two-step walking motion. The definition of knee angle is shown in $\theta$.

\begin{figure}[h]  
    \begin{center}
    \vspace{30pt}
    \includegraphics[width=70mm]{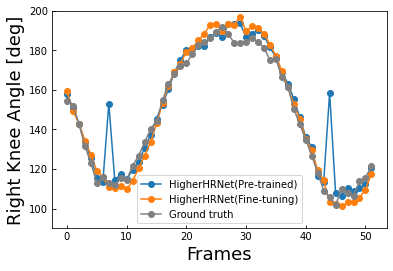}	
    \caption{Comparison of time-series changes in right knee angle to test data} 
    \label{fig:res_testdata}
    \end{center}
\end{figure}

In the pre-trained HigherHRNet, the failure of the pose estimation results in outliers in the angle change. On the other hand, the fine-tuned HigherHRNet does not have those outliers.

\newpage

\begin{figure}[h]  
    \begin{center}
    \includegraphics[width=\linewidth]{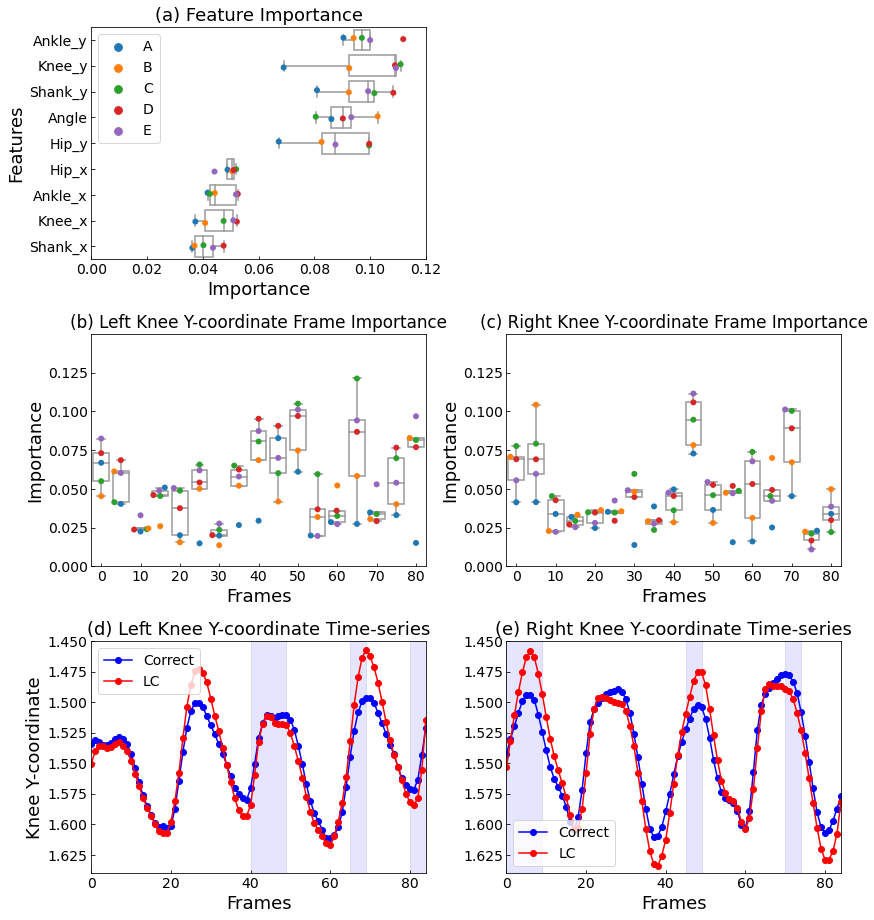}	
    \caption{Judgment grounds analysis of LC} 
    \label{fig:lc_all}
    \end{center}
\end{figure}

Figure 5(a) is the feature importance of LC judgment model. Figure 5(b) and Figure 5(c) are the Knee Y-coordinates frame importance of LC judgment model. Figure 5(d) and Figure 5(e) are the comparison of time series changes in knee Y-coordinate between normal and LC walking. The frames with relatively high feature importance are shown in light blue.

\newpage

\begin{figure}[h]  
    \begin{center}
    \includegraphics[scale=1]{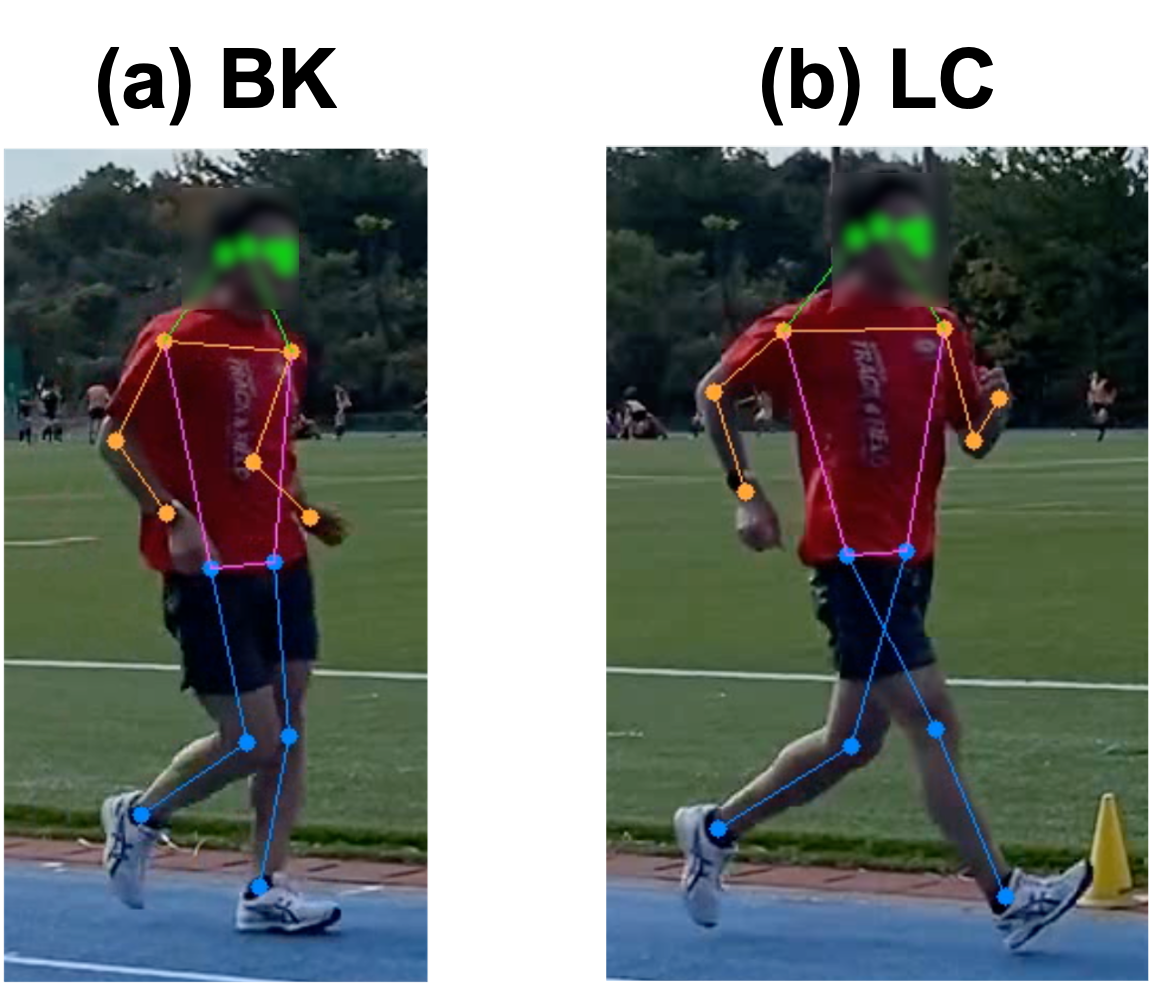}	
    \caption{The pose examples of the important frame} 
    \label{fig:bk_pose}
    \end{center}
\end{figure}

Figure 6(a) shows the pose of the important frame in the BK detection. Figure 6(b) shows the pose of the important frame in the LC detection.
\newpage

\section{Supplementary Table}

    \begin{table}[h]
        \centering
        \caption{Number of data (valid data / videos)}
        \begin{tabular}{lrrr}
            \hline
            & \multicolumn{1}{c}{Normal} & \multicolumn{1}{c}{    BK} & \multicolumn{1}{c}{    LC} \\
            \hline
            A & $31/20$ & $9/20$ & $20/20$\\
            B & $60/63$ & $94/84$ & $56/63$\\
            C & $60/60$ & $77/81$ & $64/60$\\
            D & $40/61$ & $69/80$ & $55/60$\\
            E & $64/64$ & $0/0$ & $65/65$\\
            \hline
        \end{tabular}
        \label{tab:video_num}
    \end{table}

Table 1 shows the number of videos and valid data. They are different due to failures in pose estimation and differences between the instructions at the time of walking and the judgments of the referees.

\end{document}
\end{document}